%% file: _main.tex
\documentclass{article} 
\usepackage{iclr2027_conference,times}

\input{math_commands.tex}

\usepackage{hyperref}
\usepackage{url}
\usepackage{booktabs}
\usepackage{placeins}
\usepackage{graphicx}
\usepackage{float}
\usepackage{enumitem}

\title{Cross-Lingual Alignment for Decoder-Only Models using MoE Routers}

\author{
  \begingroup
  \setlength{\tabcolsep}{4pt}
  \begin{tabular}{ccccc}
    Lucas Bandarkar & Clark Peng & Ahmed Haj Ahmed$^1$ & Aditi Khandelwal$^2$ & Nanyun Peng \\
    \multicolumn{5}{c}{\normalfont University of California, Los Angeles\thanks{Correspondence to lucasbandarkar@cs.ucla.edu}} \\
    \multicolumn{5}{c}{\normalfont \small $^1$Haverford College \quad $^2$MILA - Quebec AI Institute \& McGill University}
  \end{tabular}%
  \endgroup
}

\newcommand{\qwen}{Qwen3-30B-A3B}
\newcommand{\gpt}{GPT-OSS-20B}
\newcommand{\marco}{Marco-Nano}
\newcommand{\granite}{Granite-4.0-H-Tiny}

\iclrfinalcopy 
\begin{document}

\maketitle

\begin{abstract}
Cross-lingual contrastive learning has been a core component of multilingual encoder training, but the ability to explicitly align representations is not possible in decoder-only LLMs because of varying multilingual tokenization. However, a growing amount of research suggests that even in LLMs, higher cross-lingual representational alignment leads to improved cross-lingual transfer. In this paper, we propose a novel approach to reimagine cross-lingual contrastive learning given the architectural constraints of modern LLMs. Rather than applying an auxiliary alignment loss on hidden states, we propose using the outputs of the mixture-of-experts (MoE) routers as the target for alignment. Router outputs lend themselves better to pooling over many tokens, enabling more reliable cross-lingual comparisons at the sequence-level. 
Controlled continual pre-training experiments on four open-source MoEs show that incorporating this routing loss also aligns the underlying hidden representations across languages.
Most importantly, this loss improves multilingual performance on our diverse evaluation suite, demonstrating the potential of cross-lingual MoE router alignment.
\end{abstract}

\input{sections/intro}
\input{sections/background}
\input{sections/method}

\input{sections/experiments}
\input{sections/results}
\input{sections/discussion}
\input{sections/conclusion}



\subsubsection*{Code Availability}
Code for this work is available at \url{https://github.com/lucasbandarkar/xl_moe_alignment}.

\subsubsection*{Acknowledgments}
This research was made possible by financial support from the Amazon AI PhD Fellowship.

\bibliography{custom}
\bibliographystyle{iclr2027_conference}

\appendix
\input{appendix/method_visualization}
\input{appendix/full_results}
\input{appendix/layervis}
\input{appendix/model_layers}
\input{appendix/training_data}
\input{appendix/hyperparameters}

\end{document}

%% file: math_commands.tex
\usepackage{amsmath,amsfonts,bm}

\def\eqref#1{equation~\ref{#1}}

\def\1{\bm{1}}

\def\vp{{\bm{p}}}

\def\vz{{\bm{z}}}

\DeclareMathAlphabet{\mathsfit}{\encodingdefault}{\sfdefault}{m}{sl}
\SetMathAlphabet{\mathsfit}{bold}{\encodingdefault}{\sfdefault}{bx}{n}

\newcommand{\KL}{D_{\mathrm{KL}}}



%% file: sections/intro.tex
\begin{figure}[H]
\centering
\includegraphics[width=\linewidth]{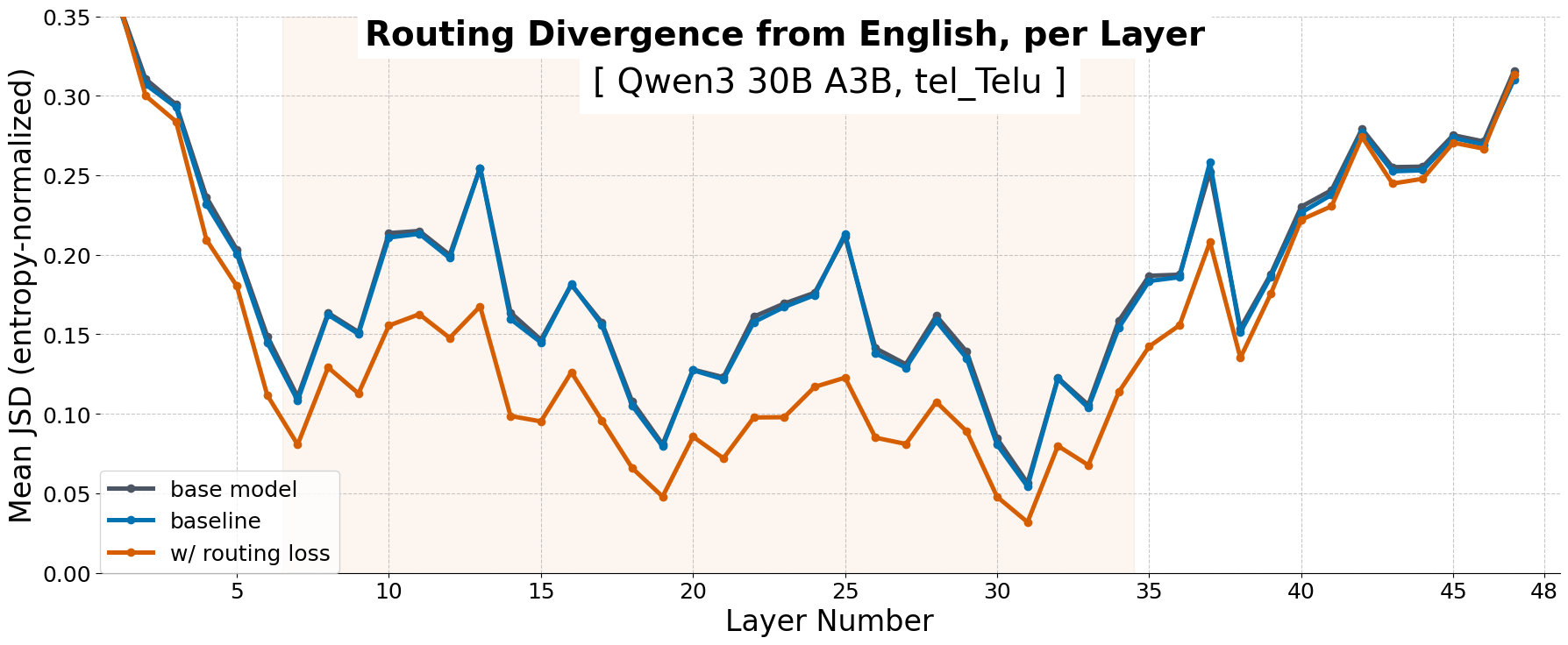}
\caption{Impact of our auxiliary alignment loss on routing alignment during CPT of the post-trained \qwen. In this case, vanilla CPT did not alter routing behavior (the blue and gray lines overlap), whereas our method (orange) substantially reduced routing divergence. The shaded orange region indicates the middle layers to which our loss was applied.}
\label{qwen}
\end{figure}

\section{Introduction}

Developing LLMs that perform well in natural languages underrepresented during pretraining or post-training remains a major challenge. One longstanding idea is that \textit{cross-lingual transfer}---the generalization of learned capabilities and knowledge across languages---is closely tied to the alignment of internal representations \citep{klementiev-etal-2012-inducing,gaschi-etal-2023-exploring,liu-niehues-2025-middle}.
If semantically equivalent inputs in different languages are represented similarly, knowledge and capabilities learned in one language are more likely to transfer to another. In earlier encoder-decoder language models, this idea could be operationalized directly because the encoder produces a natural sequence-level embedding \citep{zhou-etal-2016-cross}. Given the abundance of parallel (translated) text, cross-lingual contrastive learning could pull the embeddings of translated sentence pairs together while pushing those of unrelated pairs apart. Contrastive learning remains standard practice for training multilingual encoders \citep{wang2024multilingual,omnisonar,qwen3embedding}.

However, modern LLMs are decoder-only models. Consequently, they provide no natural sequence-level representation to align, and differences in tokenization and linguistic structure make it difficult to match tokens across translations. Common approaches to constructing sequence representations include using the final token's hidden state or mean-pooling hidden states \citep{li-etal-2024-improving-context,bridginglinguisticgaps}, but neither reliably represents the meaning of the entire sequence. As a result, multilingual LLMs are typically trained without an inductive bias toward cross-lingual representational alignment, leaving the vast supply of translation data underutilized. Even so, these models implicitly learn language-agnostic representations, particularly in their middle layers \citep{wendler-etal-2024-llamas,zhao2024multilingualism,bandarkar2026multilingual}. The extent to which models use these shared representations varies significantly across languages, and a growing body of work argues that poor cross-lingual alignment is directly tied to poor performance in those languages (See Section~\ref{sec:related_works}).


Motivated by these limitations, we present a method for effective cross-lingual alignment using the routers in mixture-of-experts (MoE) components. MoEs have become a prominent LLM architecture \citep{geminiteam2023gemini,deepseekai2025deepseekv3technicalreport} because they enable parameter scaling at a relatively fixed inference cost \citep{shazeer2017outrageously}. For each token, the router at each layer produces a probability distribution over the available experts, determining which experts to activate. Based on prior research, we hypothesize that mean-pooled routing weights provide strong alignment targets.
We evaluate this hypothesis by developing a training method that performs continual pretraining (CPT) on parallel data with an auxiliary KL-divergence loss over the mean-pooled routing weights of middle layers. We experiment with four MoE LLMs and seven languages. First, through small-scale training runs of 200k samples, we establish that this auxiliary loss effectively increases routing agreement. We also find that its gradients increase hidden-state similarity. Across a diverse set of downstream evaluations, models trained with our auxiliary loss consistently outperform those trained with vanilla CPT. Together, these results provide causal evidence that cross-lingual alignment improves multilingual performance. In practical terms, our method offers an effective and efficient cross-lingual alignment paradigm compatible with any training stage.

This paper is organized as follows. We first present recent research that motivates our method in Section~\ref{sec:related_works} and describe its design and implementation in Section~\ref{method}. Next, we outline our experiments in Section~\ref{setup} and present internal metrics and downstream results in Section~\ref{results}. Finally, we discuss our findings in Section~\ref{discussion}, and conclusions, and the opportunities they open for future work in Section~\ref{sec:conclusion}.

%% file: sections/background.tex
\section{Related Work and Hypotheses}
\label{sec:related_works}
\subsection{Cross-Lingual Representational Alignment in Encoder Models}

Because of machine translation research, there has long been a relative abundance of mined parallel data (paired multilingual texts) \citep{koehn-2005-europarl,smith-etal-2013-dirt}. Parallel data was first used to align representations across languages in early word embedding models \citep{klementiev-etal-2012-inducing}. However,languages rarely align at the word level, so later sequence-to-sequence encoder-decoder models benefitted from much more direct supervision \citep{zhou-etal-2016-cross,schwenk-douze-2017-learning,conneau-etal-2018-xnli, ouyang-etal-2021-ernie,chi-etal-2021-infoxlm}. In these models, parallel data was widely used for cross-lingual contrastive learning, which unified the feature space and facilitated multilingual generalization \citep{wu-dredze-2020-explicit,gaschi-etal-2023-exploring,patra-etal-2023-beyond}. Modern, state-of-the-art multilingual embeddings models continue to leverage such explicit contrastive signals for representation learning \citep{sturua2024jinaembeddingsv3multilingualembeddingstask, wang2024multilingual,omnisonar,qwen3embedding}. Though less clean, data augmentation can make contrastive learning also useful for monolingual models \citep{wu2020clear,Kaushik2020Learning,giorgi-etal-2021-declutr}

Highly analogous to the cross-lingual setting, contrastive learning has been a major training component of multimodal embedding models, most famously for image-text representation alignment and data-efficient vision understanding \citep{radford2021learning,zhai2023sigmoid}. However, as the field has evolved toward generative multimodal language models, vision-language integration is trained primarily with image-conditioned next-token prediction \citep{alayrac2022flamingo,liu2023visual}.

\subsection{Multilingual Representations in LLMs}

Recent work suggests that multilingual LLMs perform substantial computation in language-shared representational spaces in the middle model layers, while retaining some language-specific structure at the beginning and end \citep{kojima-etal-2024-multilingual,wu2025the,zhang2026alignment}. Recent work further shows that increasing cross-lingual alignment can improve multilingual generalization through activation patching \citep{lim2025hinders,ravisankar-etal-2026-map}, hidden-state steering \citep{lesslanguageismore,gurgurov-etal-2025-language,mahmoud2025improvingmultilinguallanguagemodels}, and router steering in MoE models \citep{bandarkar2026multilingual}. However, stronger alignment can also come at the cost of language and culture-specific information \citep{han2025rethinking}.
To improve cross-lingual alignment in decoder-only models during training, prior work has increased exposure to parallel translation data \citep{zhang-etal-2024-getting,shen-etal-2025-unaligned}.
Similar benefits from parallel data have also been observed for programming languages \citep{wu-etal-2026-parallel}. Closest to this work, \citet{liu-niehues-2025-middle} does middle layer alignment using a cosine similarity loss with hidden states.





\subsection{Central Hypotheses}

Based on these findings, our first hypothesis is, of course, that cross-lingual representation alignment causally improves cross-lingual transfer in modern LLMs.
Our second, more novel hypothesis is that using mean-pooled MoE routing weights across parallel sequences as a target can effectively align model processing across languages.
This rests on the premise that mean-pooled routing weights provide a meaningful sequence-level representation \citep{xu-etal-2026-seeing,bandarkar2026knowledge,martin2026extracting}. It has a natural interpretation: the overall expert utilization through the sequence.
By contrast, mean-pooling hidden states averages high-dimensional representation vectors, such that token-level features encoded in different directions can be attenuated or cancel under averaging.
\citet{li2025your} empirically support this distinction, showing that mean-pooled routing weights perform significantly better than mean-pooled hidden states on embedding tasks. 
Even if underlying representations remain unaligned, \citet{bandarkar2026multilingual} finds that promoting more consistent expert usage across languages is beneficial in its own right.

%% file: sections/method.tex
\section{Methodology} \label{method}

\subsection{Middle Layers Selection}

For each model, we use cross-lingual routing divergence curves following \citet{bandarkar2026multilingual}, to identify layers that already exhibit expert sharing across languages and apply the auxiliary router loss to the corresponding layers. This provides a principled way to select these layers without exhaustively searching over the combinatorially-large space of possible layer subsets.
\citet{bandarkar2026multilingual} show that these curves identify layers where cross-lingual alignment is beneficial and layers where enforcing alignment can be detrimental.
We therefore select a range of middle layers $\mathcal{M}$ for each model and hold it fixed across all experiments (see Appendix~\ref{layervis} for more details).

\subsection{Loss Formulation} \label{loss}

In order for MoE models to selectively activate a subset of experts, a \textit{router}, or gating function, takes inputs to the MoE block of a transformer decoder layer and produces logits (one for each available expert). The input hidden state are then sent to the top-$k$ experts only whose outputs are combined using normalized routing weights.
We define our notation:

\begin{itemize}[nosep,]
    \item Let $E$ be the number of experts in each MoE layer and $k << E$ be active experts per token.
    \item Let $T_{\text{eng}}, T_{\text{tgt}}$ be the sequence lengths, in tokens, of English and target-language sequences.
    \item Let $G(\cdot)$ denote the gating function that normalizes the top-$k$ logits and assigns zero weight to all remaining experts and is typically implemented as a softmax over the selected $k$ logits. 
    \item Let $\vp^{t}_{\text{eng},l}, \vp^{t}_{\text{tgt},l}$ be the routing weights for the $t^\text{th}$ token of the $i^\text{th}$ sample at layer $l$. Each $p^{t}_{\text{lang},l}$ is an $E$-dimensional probability distribution; Given router logits $\vz$, we define $\vp^{t}_{\text{lang},l} = \text{G}(\vz^{t}_{\text{lang},l})$, yielding a sparse $E$-dimensional probability distribution. 
\end{itemize}

Let $\mathcal{M}$ denote the set of layers in the selected middle-layer range. The cross-lingual routing loss is defined by first mean-pooling the routing weights across tokens in each sequence, $\bar{\vp} = \frac{1}{T}\sum_{t=1}^{T}\vp^{t}$. We then compute the KL-divergence at each layer and average these divergences across all layers in $\mathcal{M}$:
\begin{equation} \label{xlr_loss}
\mathcal{L}_{XLR} = \frac{1}{|\mathcal{M}|}\sum_{l\in\mathcal{M}} \KL (\bar{\vp}_{\text{eng},l} \Vert \bar{\vp}_{\text{tgt},l})
\end{equation}

Appendix~\ref{method_diagram} provides a visual diagram of this simple loss.

For each sample, the loss is then calculated as a weighted sum with the language-modeling loss (cross-entropy) classically used in continued pre-training:
$\mathcal{L} = \mathcal{L}_{LM} + \alpha \cdot \mathcal{L}_{XLR}$, where $\alpha$ is a hyperparameter modulating the strength of the auxiliary loss. We backpropagate gradients only through the target language sequence. The English sequence is used only to obtain routing distributions for computing $\mathcal{L}_{XLR}$, and receives no gradient updates. 

\subsection{Packing Optimizations}

\begin{figure}[h]
    \centering
    \includegraphics[width=0.99\linewidth]{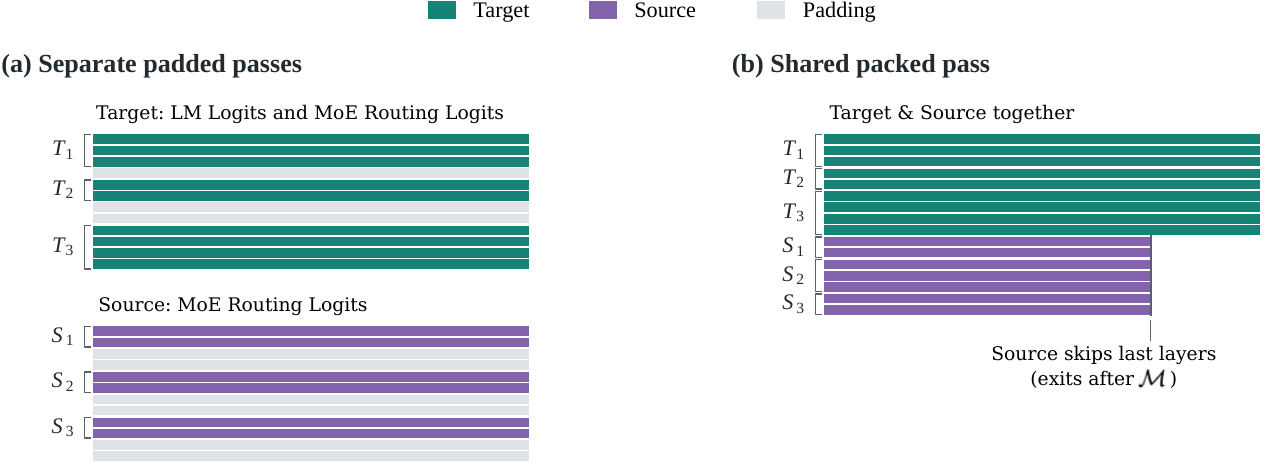}
    \caption{Comparison of the naive padded implementation (left) and our shared packed implementation (right). In the packed pass, the (typically shorter) source tokens exit early.}
    \label{fig:packing}
\end{figure}

A naive implementation requires separate forward passes for the source and target sequences to compute the routing-alignment objective, followed by an additional target-language pass for the LM objective (Figure~\ref{fig:packing}, left).
Because English has lower token fertility on average, especially compared to the non-Latin-script languages we experiment with, the second forward pass is already less expensive.
But we compute both objectives in a shared forward pass by packing source and target sequences together without padding (Figure~\ref{fig:packing}, right). Variable-length attention preserves sequence boundaries, while our split-forward implementation stops processing source tokens after the last alignment layer. 
Together, these optimizations eliminate padding and redundant computation, resulting in only a small computational overhead relative to our baseline single-language CPT.

%% file: sections/experiments.tex
\section{Experimental Setup} \label{setup}

\subsection{Models, Languages, and Training Data} \label{model_lang}

Our goal is to experiment with a diverse set of models and languages. To this end, we work with \qwen~\citep{yang2025qwen3}, \gpt~\citep{openai2025gptoss}, \granite~\citep{ibm2025granite4}, and \marco~\citep{jiang2026marcomoe}. Their MoE configurations are detailed in Appendix~\ref{details}. For the two smaller models, Granite and Marco, we experiment with Vietnamese, Sinhala, and Hungarian. For the two larger models, we use Telugu, Kannada, Thai, and Kyrgyz.
Because our experiments involve only a short CPT stage, we restrict evaluation to target languages in which the base models already exhibit non-trivial proficiency.

We construct a curated parallel training set of 200k samples for each language. We construct each training set from diverse, high-quality parallel sources. Whenever fewer curated examples are available, we supplement the training data with the OPUS translation dataset \citep{tiedemann-2012-parallel}. Indic languages are slightly over-represented (3/7) reflecting the greater availability of parallel SFT datasets for these languages. Full details are provided in Appendix~\ref{traindata}.

\subsection{Experimental Comparisons} \label{conditions}

In addition to the method described in Section~\ref{method} we implement the following for comparison:

\paragraph{Baseline.} Our baseline peforms standard CPT using only the language modeling objective $\mathcal{L}_{LM}$ applied to all tokens (not SFT). For some languages, the training data includes some parallel SFT samples, which we retain in chat-format while computing loss over the full sequence. Much of the remaining data largely consists of short, single sentence examples and therefore differs from a typical CPT distribution.
Restricting CPT to these samples nevertheless provides a controlled baseline for isolating the contribution of the auxiliary loss.


\paragraph{Router-Only Training.} We also evaluate router-only training, in which all parameters except the routers are frozen. This setting isolates the effect of directly modifying the routing function. The language modeling loss $\mathcal{L}_{LM}$ is still retained in this setting.
\citet{khandelwal2026leveraging} indicate the promise of such a parameter-efficient approach to fine-tuning.

\paragraph{Mean-Pooling Hidden States.} We additionally test our hypothesis that mean-pooling router weights is more effective than hidden states. Since hidden states are embeddings and not probability distributions, we use cosine similarity as a measure of distance between two mean-pooled sequences, as in \citet{liu-niehues-2025-middle}. Let this variant be called $\mathcal{L}_{XLH}$. We apply it to the same layers as $\mathcal{L}_{XLR}$ and all parameters are trainable. We run this training for one model, \marco, and two languages, Hungarian and Sinhala.

\subsection{Hyperparameter Selection}
Under our compute budget, we tune two hyperparameters for each model-language pair while keeping all others fixed.
The first is the learning rate, selected using baseline CPT runs for each model. The second is the $\alpha$ hyperparameter that modulates the strength of our auxiliary loss, mentioned in Section~\ref{loss},. For all models, we intitialize $\alpha$ such that $\alpha \cdot \mathcal{L}_{XLR} \approx \mathcal{L}_{LM}$ during the initial training steps, and then test nearby values. The hidden-state loss $\mathcal{L}_{XLH}$ is typically much smaller in magnitude than $\mathcal{L}_{XLR}$.
For fair comparison, we therefore use a larger $\alpha$ for $\mathcal{L}_{XLH}$, again matching the auxiliary-loss magnitude to $\approx \mathcal{L}_{LM}$.
We select hyperparameters using the validation loss (just $\mathcal{L}_{LM}$). The baseline typically achieves a lower validation loss because it optimizes only ($L_{\mathrm{LM}}$)
Final tuned values are reported in Appendix~\ref{app:complete-results} and all remaining hyperparameters that are fixed are listed in Appendix~\ref{hyperparams}.

\subsection{Evaluation}

We evaluate each checkpoint using the \verb+lm-eval-harness+ \citep{eval-harness} and a wide variety of evaluation tasks, requiring diverse latent capabilities and output formatting.
We use FLORES \citep{goyal-etal-2022-flores,nllb2022} in the eng$\rightarrow$tgt direction to evaluate target language generation and Belebele \citep{bandarkar-etal-2024-belebele} for understanding.
To evaluate cross-lingual knowledge transfer, we use the medicine subset of Global-MMLU \citep{singh-etal-2025-global} and MMLU-ProX \citep{xuan-etal-2025-mmlu}, and use MultiLoKo \citep{hupkes2025multiloko} and INCLUDE \citep{romanou2025include} for ``local'' knowledge. Global-PIQA \citep{chang2025globalpiqa} evaluates the transfer of physical reasoning. 
Finally, we evaluate mathematical reasoning with MGSM \citep{shi2023multilingual} (which includes Global-MGSM \citep{coherelabs-global-mgsm}, an extension to more languages) and PolyMath \citep{wang2025polymath}.
Because benchmark language coverage varies, the number of available tasks differs across target languages ranging from four for Kyrgyz to nine for Vietnamese.

%% file: sections/results.tex
\section{Results} \label{results}

\subsection{Impact on Routing and Representations}

\begin{figure}[!h]
\centering
\includegraphics[width=0.82\linewidth]{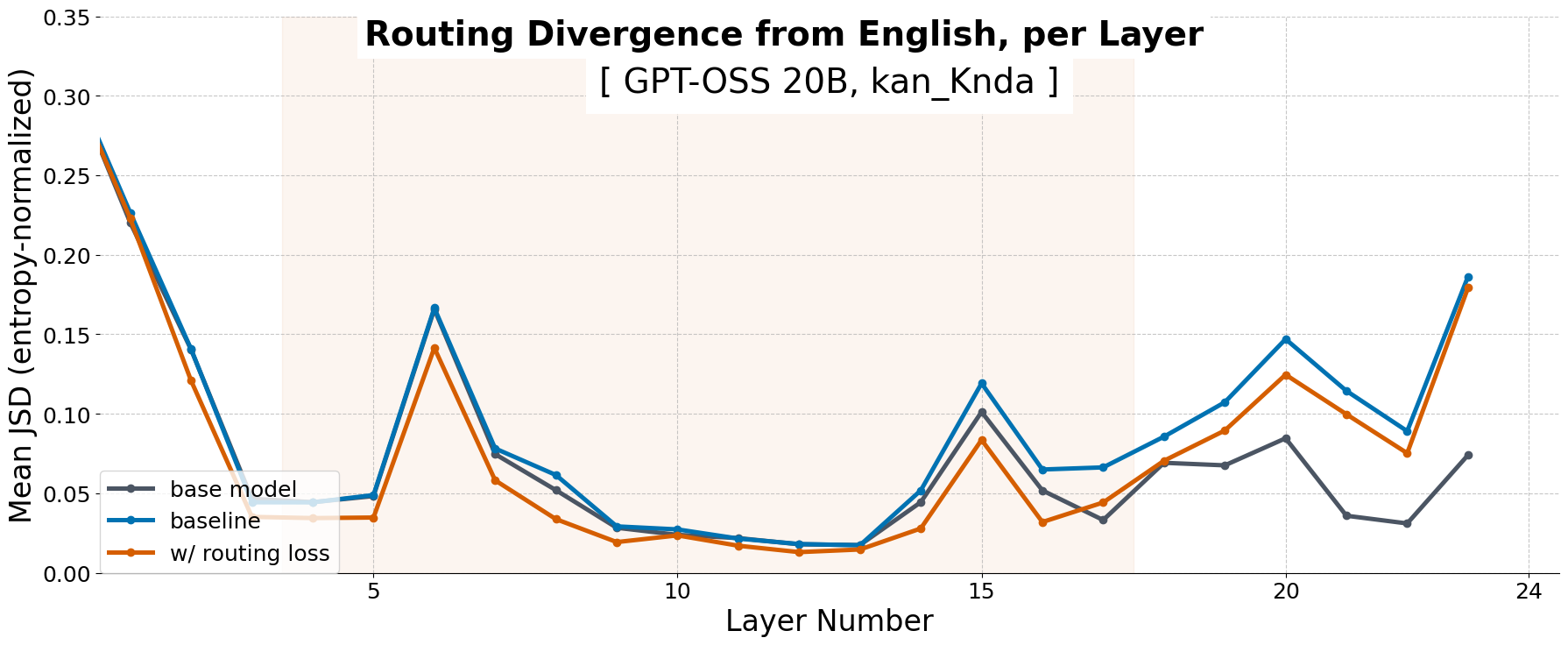}
\par\medskip
\includegraphics[width=0.82\linewidth]{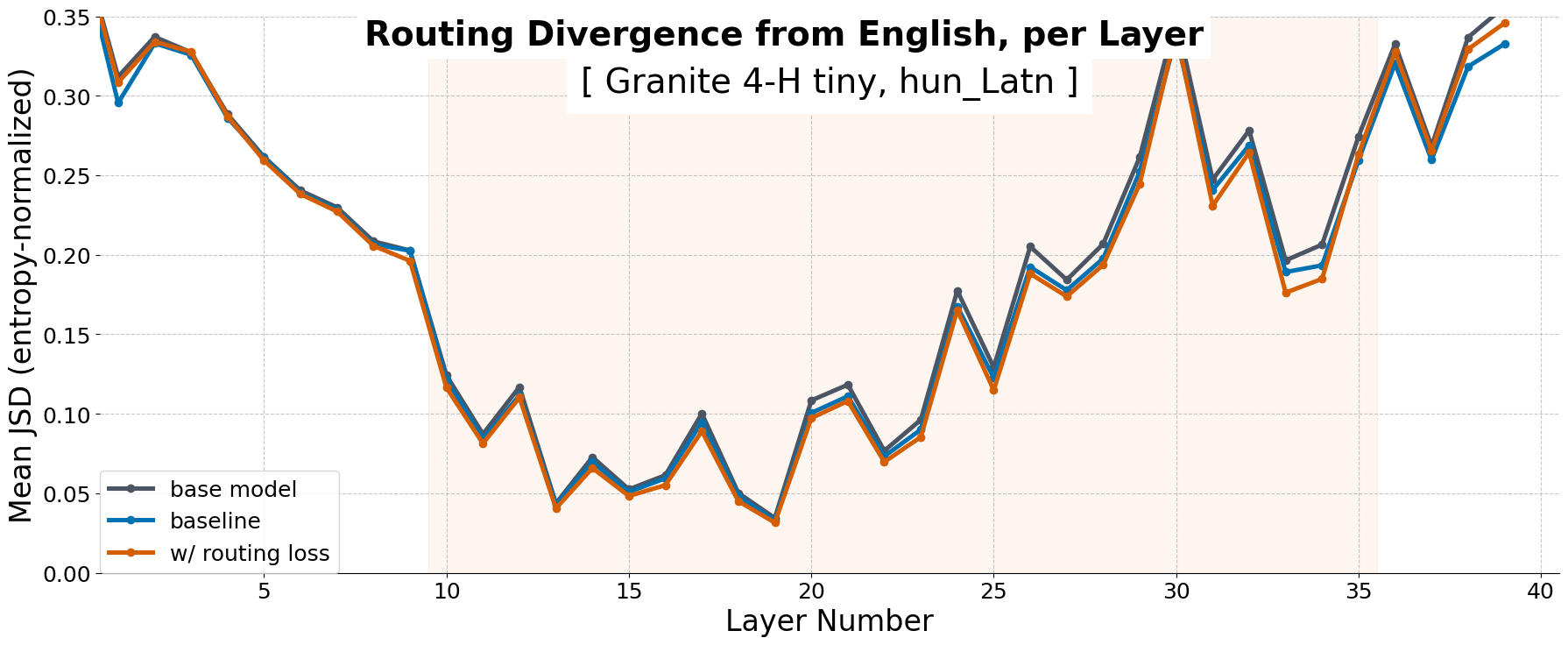}\par\medskip
\includegraphics[width=0.82\linewidth]{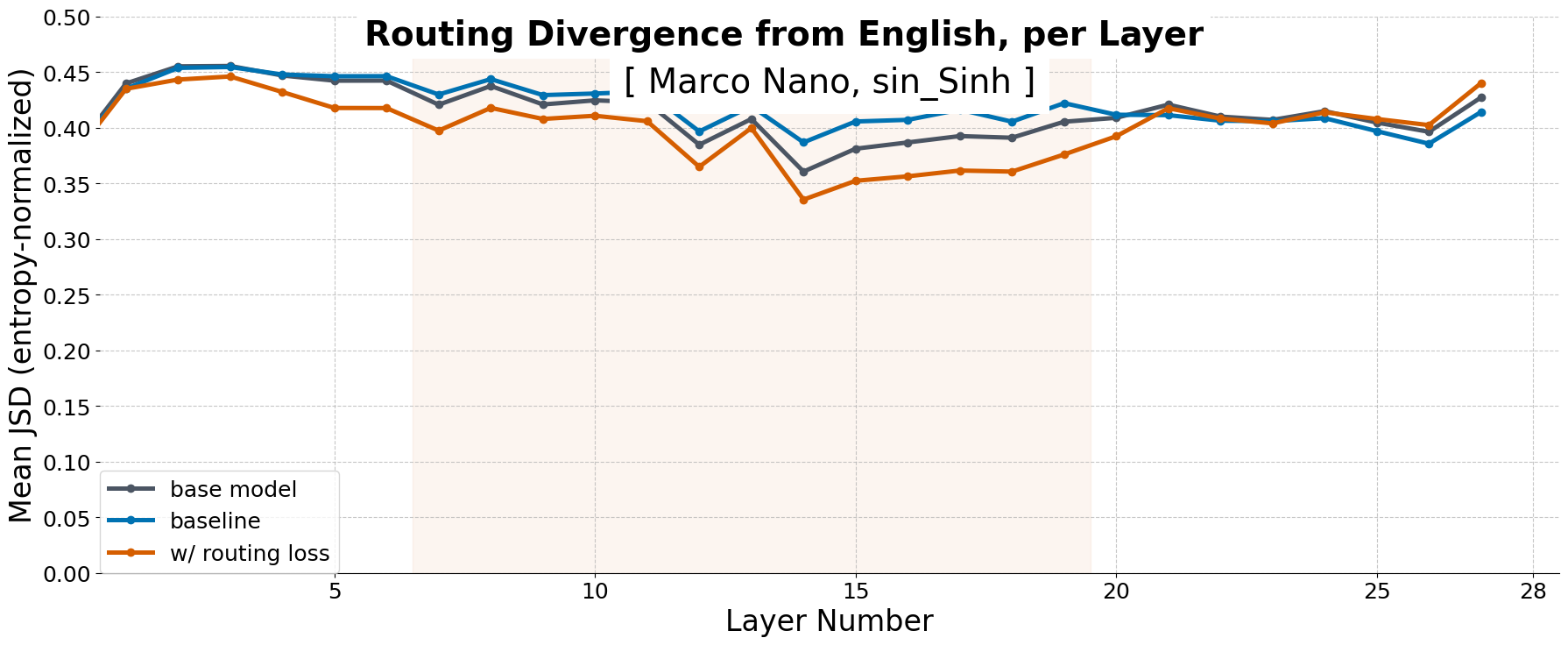}
\caption{Layer-wise routing divergence from English for \gpt{} on Kannada (top), \granite{} on Hungarian (middle), and \marco{} on Sinhala (bottom), comparing the base model, baseline training, and training with the auxiliary routing loss. Divergence is measured by mean entropy-normalized JS-divergence. The orange zone is the layers $\mathcal{M}$ where $\mathcal{L}_{XLR}$ is applied.}
\label{fig:routing-comparison}
\end{figure}

We first analyze the impact of adding our routing loss to cross-lingual alignment metrics. We use FLoRES evaluation data \citep{nllb2022}, out-of-distribution relative to any training sets, to measure this alignment. We first measure MoE routing alignment to English using the \emph{routing divergence} metric proposed by \citet{bandarkar2026multilingual}. This differs from $\mathcal{L}_{XLR}$ by using two-sided KL-divergence (JS-divergence) and normalizing values by theoretical maximum entropy for better cross-layer comparison. We visualize the difference in routing alignment for one language per model in Figures~\ref{qwen} and~\ref{fig:routing-comparison}. 
Inadvertently, the per-model hyperparameters induced varying magnitudes of changes to model routing than others. However, for all models and languages, the auxiliary loss training naturally produced the biggest increase in routing alignment (i.e. decrease in divergence). In constrast, baseline CPT led to only minor changes in routing relative to the original checkpoint.

\begin{figure}[!htbp]
    \centering
    \includegraphics[width=0.90\linewidth]{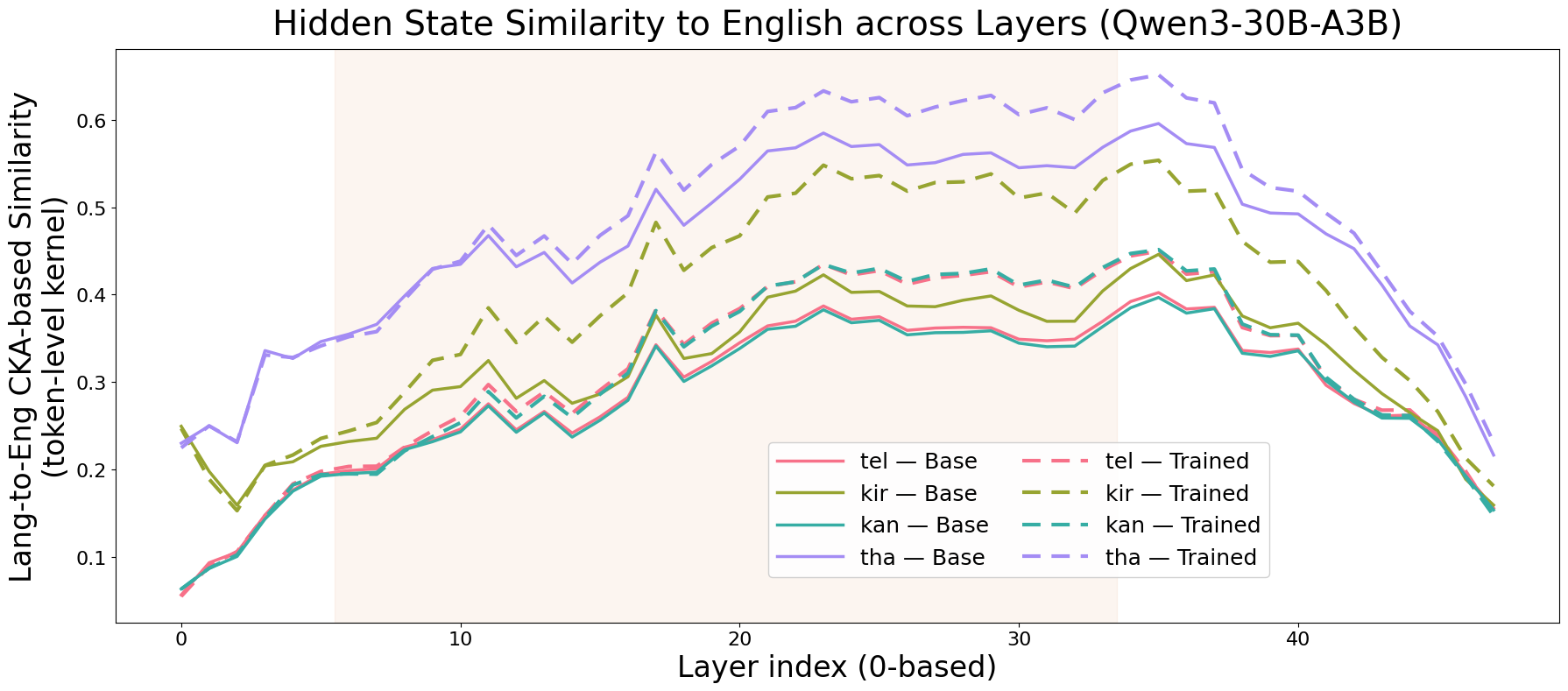}
    \caption{Layer-wise hidden-state similarity to English for Qwen3-30B-A3B. Solid lines show the base model and dashed lines show the model trained with the auxiliary cross-lingual loss for Telugu, Kyrgyz, Kannada, and Thai.}
    \label{fig:hiddenstate-xlcl-impact}
\end{figure}

MoE router alignment does not directly imply hidden state representation alignment. But Gradients from the routing loss propagate through preceding model parameters. We therefore evaluate whether the routing loss also changes the underlying hidden state representations, rather than only the router outputs. Moreover, activating similar experts may increase representational similarity in subsequent layers. To measure the impact on underlying representations, we perform forward passes on the same parallel data and collect the hidden states at two points each decoder layer: entering the layer and the hidden state after the attention block\footnote{Concretely, this is the hidden state entering the MoE block, right after the residual connection is re-added.}. Then, we calculate the SoftCKA cross-lingual alignment metric from \citet{bandarkar2026multilingualityhybridattentionllms}. Rather than mean-pooling, this calculates within-sequence and cross-language RBF kernel matrices to soft-match tokens. This allows us to use centered kernel alignment (CKA) \citep{pmlr-v97-kornblith19a} without explicitly matching tokens across sequences. This metric is displayed for Qwen3 in Figure~\ref{fig:hiddenstate-xlcl-impact}. It clearly displays that across languages, our MoE alignment metric is making representations, themselves, more similar to English.


\input{tables/results_summary}

\subsection{Impact on Downstream Tasks}

\input{tables/hidden_state_comparison}

Because benchmark coverage differs by language, Table~\ref{tab:results-summary} reports the average over all available tasks for each model--language pair; task-level scores and complete training configurations are provided in Appendix~\ref{app:complete-results}.
The baseline itself yields only modest gains over the original checkpoint, so differences between experimental conditions are also relatively small.
At the individual task level, there is noise. However, the language-level averages show a consistent trend. The routing loss improves over this baseline for 13 of the 14 model--language pairs and ties for the remaining pair (\marco{} on Vietnamese). The positive gains range up to 2.1 average points, with a mean improvement of 0.9 points across all pairs.
This gains are observed despite the relatively short CPT stage applied earlier to already post-trained checkpoints.

Table~\ref{tab:hidden-state-comparison} summarizes the comparison to mean-pooling hidden states using cosine similarity ($\mathcal{L}_{XLH}$). The hidden-state objective is competitive on Sinhala, improving the baseline average from 26.1 to 26.3, but remains below the 26.8 obtained with $\mathcal{L}_{XLR}$. On Hungarian, hidden-state alignment actually decreases the baseline score from 37.4 to 36.9, whereas routing alignment increases it to 38.1. Thus, in both tested settings, mean-pooled router distributions provide a more effective alignment target than mean-pooled hidden representations. Because this ablation covers only \marco{} and two languages, we view it as targeted support for our design choice rather than a comprehensive comparison of representation-alignment methods.

We next analyze the router-only training condition.
Despite updating fewer than $<0.1\%$ of params, it ocasionally matches or exceeds the LM-only baseline, although performance varies substantially across settings.
Its changes to the internal alignment metrics are similarly noisy.
These results suggest that router-only training may provide a parameter-efficient alternative, although it is not a reliable substitute for full-model training.
Taken together, these results suggest that the gains from the routing loss arise primarily from changes to the representations supplied to the routers, rather than the updates to the routing function alone.
\FloatBarrier

%% file: tables/results_summary.tex
\begin{table}[!t]
\centering
\footnotesize
\setlength{\tabcolsep}{3pt}
\renewcommand{\arraystretch}{1.0}
\caption{Average benchmark scores for each model, language, and training condition. Parentheses in the language headers indicate the number of tasks included in each average. Bold marks the best score within each model--language pair, including ties at the reported precision. Original checkpoint denotes the released post-trained model before our training. The baseline uses only target-language LM training; our method adds the routing-alignment loss. Router-only training applies the auxiliary loss while updating only the routers in the selected middle layers.}
\label{tab:results-summary}
\begin{tabular}{@{}llcccc@{}}
\toprule
Model & Condition & Telugu (8) & Kyrgyz (4) & Kannada (4) & Thai (7) \\
\midrule
\qwen{} & Original checkpoint & 44.6 & 40.8 & 49.3 & 43.0 \\
 & Baseline & 45.4 & 43.5 & 49.6 & 44.9 \\
 & + aux routing loss (ours) & \textbf{46.3} & \textbf{45.6} & \textbf{51.4} & \textbf{45.4} \\
 & Router-only training & 45.1 & 42.2 & 49.9 & 43.8 \\
\addlinespace
\gpt{} & Original checkpoint & 42.5 & 35.9 & 47.5 & 34.9 \\
 & Baseline & 46.5 & 42.0 & 55.5 & 39.9 \\
 & + aux routing loss (ours) & \textbf{47.0} & \textbf{43.8} & \textbf{56.4} & \textbf{40.8} \\
 & Router-only training & 43.2 & 36.4 & 48.8 & 34.7 \\
\midrule
Model & Condition & Vietnamese (9) & Sinhala (5) & Hungarian (6) & \\
\midrule
\granite{} & Original checkpoint & 34.8 & 31.3 & 34.5 &  \\
 & Baseline & 36.5 & 32.8 & 38.1 &  \\
 & + aux routing loss (ours) & \textbf{36.8} & \textbf{32.9} & \textbf{38.8} &  \\
 & Router-only training & 36.5 & 31.9 & 38.2 &  \\
\addlinespace
\marco{} & Original checkpoint & 44.2 & 24.0 & 36.3 &  \\
 & Baseline & \textbf{45.6} & 26.1 & 37.4 &  \\
 & + aux routing loss (ours) & \textbf{45.6} & \textbf{26.8} & \textbf{38.1} &  \\
 & Router-only training & 44.8 & 25.2 & 37.1 &  \\
\bottomrule
\end{tabular}
\end{table}

%% file: tables/hidden_state_comparison.tex
\begin{table}[!t]
\centering
\footnotesize
\renewcommand{\arraystretch}{1.0}
\caption{Routing versus hidden-state alignment for \marco{}. Entries are the reported average benchmark scores over the benchmarks available in each language; parentheses in the headers give the number of tasks in each average. Bold marks the best score in each column. Hidden-state alignment uses $\alpha=5000$; routing alignment uses $\alpha=10$ for both languages. Full hyperparameters and task-level scores appear in Table~\ref{tab:results-marco}.}
\label{tab:hidden-state-comparison}
\begin{tabular}{@{}lrr@{}}
\toprule
Condition & Sinhala (5) & Hungarian (6) \\
\midrule
Original checkpoint & 24.0 & 36.3 \\
Baseline & 26.1 & 37.4 \\
+ routing loss (ours) & \textbf{26.8} & \textbf{38.1} \\
+ hidden-state loss & 26.3 & 36.9 \\
\bottomrule
\end{tabular}
\end{table}

%% file: sections/discussion.tex
\section{Discussion and Future Work} \label{discussion}


\paragraph{Experimental Limitations} In order to evaluate the impact of our training method on downstream tasks, we were constrained to experiment on fully post-trained LLMs. Using base models would lead to noisy evaluations, given they would not have learned to reason, perform chain-of-thought, follow instructions, etc. On the other hand, the abundant parallel data available is largely unlabeled, requiring us to do CPT. CPT on post-trained LLMs is sensitive and inefficient, as is reflected in the very limited gains by the baseline. We elect such an experimental setting to control for the influence of the routing loss on downstream evaluations, likely at the expense of bigger experimental deltas.

\paragraph{Real-Life Implementation}
We believe this method would be much more effective during pretraining rather than post-training because that would compound all the learnings from pre-, mid-, and post-training to be tied together across languages more natively. 
In addition, recent studies suggest MoE routing dynamics are established quite early on in pretraining \citep{xue2024openmoe,muennighoff2025olmoe}.

\paragraph{Dependence on Parallel Data} Naturally, this method is entirely reliant on parallel data. This restricts its applicability because much of the data used for pre-training, mid-training, SFT or RL is not parallel. 
Existing parallel corpora are also predominantly sentence-level and cover a relatively narrow data distribution. As can be seen in our baseline, where the model only improved a small amount during a CPT of 200k training samples, this data on its own cannot substantially improve a model on downstream tasks. We therefore view MoE alignment as one component of a broader training curriculum, interleaved with stages that optimize on complementary data distributions.


\paragraph{No Negatives} Often, the term \emph{contrastive} learning implies the presence of positive and negative pairs. The purpose of the negative pairs is to prevent representation collapse \citep{infonce}. However, the negative pairs are unnecessary here as MoE LLMs are always trained with load-balancing (another auxiliary loss), which incentivizes equitable use of experts across different batch sizes. In our experiments, our alignment loss is only done at a small-scale, with low risk of collapsing the router distributions. As a result, we do not experiment with negative pairs nor with the interaction with load-balancing auxiliary loss. We leave this to future work, as such experiments would likely require large-scale (i.e. expensive) pretraining.

\paragraph{Transfer-Dependent Tasks}
Based on prior work \citep{zhu-etal-2024-question,yoon-etal-2024-langbridge,hu-etal-2025-large-language}, we additionally hypothesize that alignment may be more beneficial for tasks that depend less strongly on language-specific information, like reasoning, math, knowledge, etc.
However, our results do not establish convincingly that router alignment benefits such tasks more than translation and reading comprehension.


\paragraph{Direct Early-Layer Gradients} In traditional CPT, the LM loss comes from the logit distribution, which is subject to a gradient bottleneck during backpropagation through the softmax operation \citep{godey2026lost}. Auxiliary internal losses such as MoE load-balancing and our routing divergence loss, provide more direct supervision to early-layer parameters helping mitigate vanishing or exploding gradients \citep{lee2015deeply,kedia2024transformers,li2025mixln}.

\paragraph{Cross-Modal Representation Learning} While contrastive learning has been standard practice in encoder-based architectures \citep{radford2021learning}, modern multimodal generative models do not use such supervision \citep{wang2024emu3}. Recent work has further identified modality-dependent routing discrepancies in multimodal MoE models that can hinder the activation of task-relevant reasoning experts \citep{xu-etal-2026-seeing}. Future work could investigate whether MoE-based alignment could facilitate cross-modal alignment and improve reasoning capabilities across modalities.

%% file: sections/conclusion.tex
\section{Conclusion}
\label{sec:conclusion}
This work introduces a cross-lingual alignment objective for decoder-only MoE language models that minimizes the divergence between mean-pooled routing distributions for translated sequences. Grounded in recent LLM research on multilinguality, our method provides a pragmatic way to introduce sequence-level cross-lingual supervision into modern decoder-only architectures without requiring token-level correspondence. We find that its impact extends beyond the explicit routing objective by also aligning the model's hidden representations. Consistent with the hypothesis that representational alignment facilitates cross-lingual transfer, these internal changes yield positive aggregate gains across diverse multilingual evaluations, even within our limited-scale post-training setting. Together, these results provide empirical support for the role of cross-lingual alignment in multilingual transfer and demonstrate that it can be directly encouraged in MoE LLMs.


These findings establish MoE routing as a promising interface for inducing cross-lingual transfer in modern LLMs. Future work could develop objectives that reduce token-level information loss, determine when and where alignment supervision is most effective within a model's training curriculum, and reduce reliance on explicitly parallel data. Studying these questions at pretraining scale may reveal more general ways to encourage shared representations without sacrificing language- or modality-specific structure.

%% file: appendix/method_visualization.tex
\section{Visual Diagram of Our Methodology} \label{method_diagram}

\begin{figure}[H]
    \centering
    \includegraphics[width=\textwidth]{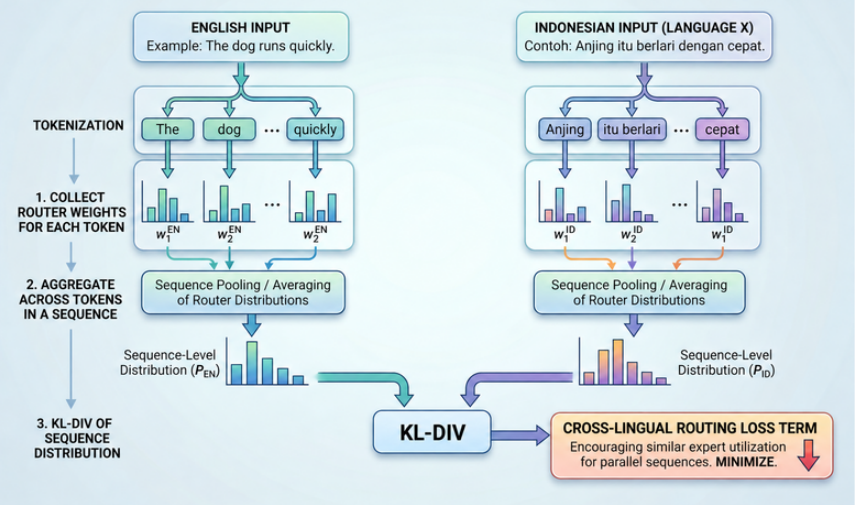}
    \caption{Overview of our cross-lingual alignment method to supplement the formal description in Section~\ref{loss}.}
    \label{fig:method_diagram}
\end{figure}

%% file: appendix/full_results.tex
\section{Complete Experimental Results}
\label{app:complete-results}

Tables~\ref{tab:results-qwen3}--\ref{tab:results-marco} use one row per experimental condition. \emph{Original} denotes the released post-trained checkpoint before our training, not a base model. Baseline is target-language LM-only training; + routing adds our auxiliary loss; Router-only updates only the selected routers; + hidden aligns hidden states instead. $\alpha$ is the auxiliary-loss coefficient. LM loss excludes auxiliary losses; AVG averages the available benchmarks. Percentage signs are omitted, and FLORES retains its original scale. MMLU Med. and PIQA denote Global MMLU Medical and Global PIQA. Dashes indicate unavailable or inapplicable values; \texttt{scaled\_alpha} is omitted.

\input{tables/results_qwen3}
\clearpage
\input{tables/results_gpt}
\clearpage
\input{tables/results_granite}
\clearpage
\input{tables/results_marco}

%% file: tables/results_qwen3.tex
\begin{table}[!htbp]
\centering
\setlength{\tabcolsep}{2pt}
\renewcommand{\arraystretch}{1}
\caption{Complete results for \qwen{}. Auxiliary alignment is applied to layers 7--34. Bold AVG values mark the best result per language, including ties.}
\label{tab:results-qwen3}
\begin{tabular*}{\linewidth}{@{\extracolsep{\fill}}llrrrr@{}}
\toprule
\multicolumn{6}{@{}l}{\textbf{(a) Training configuration and aggregate results}} \\
\midrule
Language & Condition & $\alpha$ & Learning rate & LM loss & AVG \\
\midrule
Telugu (tel) & Original & -- & -- & 0.77 & 44.6 \\
 & Baseline & -- & $3\times 10^{-6}$ & 0.65 & 45.4 \\
 & \textbf{+ routing} & 2 & $3\times 10^{-6}$ & 0.7 & \textbf{46.3} \\
 & Router-only & 10 & $3\times 10^{-6}$ & 0.78 & 45.1 \\
\addlinespace[3pt]
Kyrgyz (kir) & Original & -- & -- & 3.06 & 40.8 \\
 & Baseline & -- & $4\times 10^{-6}$ & 1.95 & 43.5 \\
 & \textbf{+ routing} & 10 & $4\times 10^{-6}$ & 2.07 & \textbf{45.6} \\
 & Router-only & 10 & $4\times 10^{-6}$ & 3 & 42.2 \\
\addlinespace[3pt]
Kannada (kan) & Original & -- & -- & 0.81 & 49.3 \\
 & Baseline & -- & $3.5\times 10^{-6}$ & 0.69 & 49.6 \\
 & \textbf{+ routing} & 4 & $2.5\times 10^{-6}$ & 0.76 & \textbf{51.4} \\
 & Router-only & 6 & $3\times 10^{-6}$ & 0.81 & 49.9 \\
\addlinespace[3pt]
Thai (tha) & Original & -- & -- & 2.63 & 43.0 \\
 & Baseline & -- & $3\times 10^{-6}$ & 2.2 & 44.9 \\
 & \textbf{+ routing} & 10 & $3\times 10^{-6}$ & 2.36 & \textbf{45.4} \\
 & Router-only & 10 & $4\times 10^{-6}$ & 2.59 & 43.8 \\
\bottomrule
\end{tabular*}
\par\vspace{1em}
\begin{tabular*}{\linewidth}{@{\extracolsep{\fill}}llrrrrrrrrr@{}}
\toprule
\multicolumn{11}{@{}l}{\textbf{(b) Benchmark scores}} \\
\midrule
Lang. & Condition & Belebele & FLORES & \shortstack{MMLU\\Med.} & PIQA & INCLUDE & MGSM & \shortstack{MMLU\\Pro-X} & \shortstack{Multi\\LoKo} & \shortstack{Poly\\Math} \\
\midrule
tel & Original & 70.7 & 32.8 & 60.2 & 68.0 & 50.4 & 10.0 & 25.8 & -- & 39.2 \\
 & Baseline & 68.0 & 31.6 & 60.0 & 69.0 & 49.5 & 20.8 & 24.5 & -- & 40.2 \\
 & \textbf{+ routing} & 72.8 & 29.0 & 61.7 & 69.0 & 50.0 & 22.4 & 22.1 & -- & 43.2 \\
 & Router-only & 71.0 & 33.4 & 61.9 & 66.0 & 48.9 & 15.6 & 24.5 & -- & 39.8 \\
\addlinespace[3pt]
kir & Original & 69.6 & 18.3 & -- & 57.0 & -- & 18.4 & -- & -- & -- \\
 & Baseline & 70.9 & 21.2 & -- & 58.0 & -- & 24.0 & -- & -- & -- \\
 & \textbf{+ routing} & 72.9 & 20.5 & -- & 62.0 & -- & 26.8 & -- & -- & -- \\
 & Router-only & 69.1 & 18.8 & -- & 60.0 & -- & 20.8 & -- & -- & -- \\
\addlinespace[3pt]
kan & Original & 75.2 & 32.4 & -- & 63.0 & -- & 26.4 & -- & -- & -- \\
 & Baseline & 74.8 & 31.6 & -- & 65.0 & -- & 26.8 & -- & -- & -- \\
 & \textbf{+ routing} & 78.9 & 29.4 & -- & 63.0 & -- & 34.4 & -- & -- & -- \\
 & Router-only & 76.3 & 33.0 & -- & 64.0 & -- & 26.4 & -- & -- & -- \\
\addlinespace[3pt]
tha & Original & 83.6 & 24.8 & -- & 69.0 & -- & 11.6 & 58.0 & 12.8 & 41.2 \\
 & Baseline & 83.7 & 27.7 & -- & 68.0 & -- & 30.4 & 49.2 & 14.0 & 41.6 \\
 & \textbf{+ routing} & 84.4 & 24.5 & -- & 66.0 & -- & 33.6 & 51.2 & 14.4 & 43.8 \\
 & Router-only & 83.9 & 24.5 & -- & 69.0 & -- & 16.0 & 59.2 & 12.8 & 41.2 \\
\bottomrule
\end{tabular*}
\end{table}

%% file: tables/results_gpt.tex
\begin{table}[!htbp]
\centering
\setlength{\tabcolsep}{2pt}
\renewcommand{\arraystretch}{1}
\caption{Complete results for \gpt{}. Auxiliary alignment is applied to layers 4--17. Bold AVG values mark the best result per language, including ties.}
\label{tab:results-gpt}
\begin{tabular*}{\linewidth}{@{\extracolsep{\fill}}llrrrr@{}}
\toprule
\multicolumn{6}{@{}l}{\textbf{(a) Training configuration and aggregate results}} \\
\midrule
Language & Condition & $\alpha$ & Learning rate & LM loss & AVG \\
\midrule
Telugu (tel) & Original & -- & -- & 3.05 & 42.5 \\
 & Baseline & -- & $4\times 10^{-7}$ & 2.69 & 46.5 \\
 & \textbf{+ routing} & 40 & $4\times 10^{-7}$ & 2.87 & \textbf{47.0} \\
 & Router-only & 40 & $8\times 10^{-7}$ & 3.04 & 43.2 \\
\addlinespace[3pt]
Kyrgyz (kir) & Original & -- & -- & 4.28 & 35.9 \\
 & Baseline & -- & $6\times 10^{-7}$ & 3.4 & 42.0 \\
 & \textbf{+ routing} & 20 & $4\times 10^{-7}$ & 3.64 & \textbf{43.8} \\
 & Router-only & 40 & $5\times 10^{-7}$ & 4.27 & 36.4 \\
\addlinespace[3pt]
Kannada (kan) & Original & -- & -- & 2.99 & 47.5 \\
 & Baseline & -- & $6\times 10^{-7}$ & 2.56 & 55.5 \\
 & \textbf{+ routing} & 40 & $6\times 10^{-7}$ & 2.76 & \textbf{56.4} \\
 & Router-only & 40 & $8\times 10^{-7}$ & 2.96 & 48.8 \\
\addlinespace[3pt]
Thai (tha) & Original & -- & -- & 3.69 & 34.9 \\
 & Baseline & -- & $6\times 10^{-7}$ & 3.18 & 39.9 \\
 & \textbf{+ routing} & 40 & $8\times 10^{-7}$ & 3.47 & \textbf{40.8} \\
 & Router-only & 20 & $1\times 10^{-6}$ & 3.67 & 34.7 \\
\bottomrule
\end{tabular*}
\par\vspace{1em}
\begin{tabular*}{\linewidth}{@{\extracolsep{\fill}}llrrrrrrrrr@{}}
\toprule
\multicolumn{11}{@{}l}{\textbf{(b) Benchmark scores}} \\
\midrule
Lang. & Condition & Belebele & FLORES & \shortstack{MMLU\\Med.} & PIQA & INCLUDE & MGSM & \shortstack{MMLU\\Pro-X} & \shortstack{Multi\\LoKo} & \shortstack{Poly\\Math} \\
\midrule
tel & Original & 58.3 & 28.3 & 59.8 & 69.0 & 37.4 & 18.8 & 35.2 & -- & 32.8 \\
 & Baseline & 69.4 & 39.2 & 61.7 & 68.0 & 38.7 & 25.6 & 30.2 & -- & 39.4 \\
 & \textbf{+ routing} & 68.9 & 41.3 & 61.0 & 69.0 & 38.5 & 26.0 & 28.4 & -- & 42.8 \\
 & Router-only & 59.3 & 27.9 & 59.5 & 68.0 & 37.0 & 20.4 & 36.9 & -- & 36.4 \\
\addlinespace[3pt]
kir & Original & 31.2 & 18.1 & -- & 62.0 & -- & 32.4 & -- & -- & -- \\
 & Baseline & 50.2 & 19.1 & -- & 62.0 & -- & 36.8 & -- & -- & -- \\
 & \textbf{+ routing} & 44.2 & 19.1 & -- & 67.0 & -- & 44.8 & -- & -- & -- \\
 & Router-only & 31.4 & 18.2 & -- & 60.0 & -- & 36.0 & -- & -- & -- \\
\addlinespace[3pt]
kan & Original & 62.2 & 26.2 & -- & 63.0 & -- & 38.6 & -- & -- & -- \\
 & Baseline & 72.0 & 40.0 & -- & 63.0 & -- & 46.8 & -- & -- & -- \\
 & \textbf{+ routing} & 70.9 & 41.8 & -- & 65.0 & -- & 48.0 & -- & -- & -- \\
 & Router-only & 61.8 & 26.6 & -- & 64.0 & -- & 42.8 & -- & -- & -- \\
\addlinespace[3pt]
tha & Original & 50.9 & 25.9 & -- & 65.0 & -- & 15.2 & 50.2 & 1.6 & 35.6 \\
 & Baseline & 70.2 & 37.3 & -- & 66.0 & -- & 14.0 & 47.5 & 8.4 & 36.2 \\
 & \textbf{+ routing} & 70.6 & 35.2 & -- & 67.0 & -- & 19.6 & 46.0 & 8.2 & 39.2 \\
 & Router-only & 50.9 & 27.5 & -- & 64.0 & -- & 14.0 & 48.0 & 2.4 & 36.2 \\
\bottomrule
\end{tabular*}
\end{table}

%% file: tables/results_granite.tex
\begin{table}[!htbp]
\centering
\setlength{\tabcolsep}{2pt}
\renewcommand{\arraystretch}{1}
\caption{Complete results for \granite{}. Auxiliary alignment is applied to layers 10--35. Bold AVG values mark the best result per language, including ties.}
\label{tab:results-granite}
\begin{tabular*}{\linewidth}{@{\extracolsep{\fill}}llrrrr@{}}
\toprule
\multicolumn{6}{@{}l}{\textbf{(a) Training configuration and aggregate results}} \\
\midrule
Language & Condition & $\alpha$ & Learning rate & LM loss & AVG \\
\midrule
Vietnamese (vie) & Original & -- & -- & 2.93 & 34.8 \\
 & Baseline & -- & $4\times 10^{-7}$ & 2.31 & 36.5 \\
 & \textbf{+ routing} & 200 & $5\times 10^{-7}$ & 2.402 & \textbf{36.8} \\
 & Router-only & 200 & $4\times 10^{-7}$ & 2.85 & 36.5 \\
\addlinespace[3pt]
Sinhala (sin) & Original & -- & -- & 1.09 & 31.3 \\
 & Baseline & -- & $8\times 10^{-7}$ & 0.95 & 32.8 \\
 & \textbf{+ routing} & 200 & $6\times 10^{-7}$ & 0.99 & \textbf{32.9} \\
 & Router-only & 100 & $4\times 10^{-7}$ & 1.09 & 31.9 \\
\addlinespace[3pt]
Hungarian (hun) & Original & -- & -- & 4.77 & 34.5 \\
 & Baseline & -- & $8\times 10^{-7}$ & 3.38 & 38.1 \\
 & \textbf{+ routing} & 50 & $8\times 10^{-7}$ & 3.61 & \textbf{38.8} \\
 & Router-only & 100 & $4\times 10^{-7}$ & 4.49 & 38.2 \\
\bottomrule
\end{tabular*}
\par\vspace{1em}
\begin{tabular*}{\linewidth}{@{\extracolsep{\fill}}llrrrrrrrrr@{}}
\toprule
\multicolumn{11}{@{}l}{\textbf{(b) Benchmark scores}} \\
\midrule
Lang. & Condition & Belebele & FLORES & \shortstack{MMLU\\Med.} & PIQA & INCLUDE & MGSM & \shortstack{MMLU\\Pro-X} & \shortstack{Multi\\LoKo} & \shortstack{Poly\\Math} \\
\midrule
vie & Original & 62.6 & 37.7 & 47.4 & 59.0 & 46.9 & 23.2 & 9.1 & 8.0 & 18.8 \\
 & Baseline & 64.2 & 38.3 & 48.6 & 60.0 & 46.4 & 31.6 & 10.2 & 9.6 & 19.6 \\
 & \textbf{+ routing} & 65.0 & 38.0 & 48.1 & 62.0 & 46.0 & 32.8 & 10.8 & 9.2 & 19.0 \\
 & Router-only & 64.1 & 39.2 & 48.8 & 60.0 & 46.7 & 32.4 & 8.8 & 9.6 & 18.6 \\
\addlinespace[3pt]
sin & Original & 47.6 & 14.6 & 42.9 & 48.0 & -- & 3.2 & -- & -- & -- \\
 & Baseline & 47.0 & 16.2 & 43.6 & 51.0 & -- & 6.0 & -- & -- & -- \\
 & \textbf{+ routing} & 47.4 & 16.8 & 44.5 & 50.0 & -- & 5.6 & -- & -- & -- \\
 & Router-only & 47.7 & 16.8 & 43.6 & 48.0 & -- & 3.6 & -- & -- & -- \\
\addlinespace[3pt]
hun & Original & 61.8 & 25.1 & -- & 60.0 & 40.5 & 16.0 & 3.7 & -- & -- \\
 & Baseline & 60.3 & 25.0 & -- & 59.0 & 41.8 & 37.6 & 5.0 & -- & -- \\
 & \textbf{+ routing} & 60.7 & 25.9 & -- & 60.0 & 40.5 & 40.0 & 5.8 & -- & -- \\
 & Router-only & 59.9 & 26.1 & -- & 59.0 & 41.6 & 38.0 & 4.8 & -- & -- \\
\bottomrule
\end{tabular*}
\end{table}

%% file: tables/results_marco.tex
\begin{table}[!htbp]
\centering
\setlength{\tabcolsep}{2pt}
\renewcommand{\arraystretch}{1}
\caption{Complete results for \marco{}. Auxiliary alignment is applied to layers 7--19. Bold AVG values mark the best result per language, including ties. Hidden-state alignment is evaluated only for Sinhala and Hungarian.}
\label{tab:results-marco}
\begin{tabular*}{\linewidth}{@{\extracolsep{\fill}}llrrrr@{}}
\toprule
\multicolumn{6}{@{}l}{\textbf{(a) Training configuration and aggregate results}} \\
\midrule
Language & Condition & $\alpha$ & Learning rate & LM loss & AVG \\
\midrule
Vietnamese (vie) & Original & -- & -- & 3.46 & 44.2 \\
 & Baseline & -- & $1\times 10^{-6}$ & 3.16 & \textbf{45.6} \\
 & \textbf{+ routing} & 10 & $1.2\times 10^{-6}$ & 3.23 & \textbf{45.6} \\
 & Router-only & 40 & $2\times 10^{-6}$ & 3.45 & 44.8 \\
\addlinespace[3pt]
Sinhala (sin) & Original & -- & -- & 1.79 & 24.0 \\
 & Baseline & -- & $1\times 10^{-6}$ & 1.57 & 26.1 \\
 & \textbf{+ routing} & 10 & $1.3\times 10^{-6}$ & 1.59 & \textbf{26.8} \\
 & Router-only & 20 & $1.2\times 10^{-6}$ & 1.79 & 25.2 \\
 & + hidden & 5000 & $1.2\times 10^{-6}$ & 1.74 & 26.3 \\
\addlinespace[3pt]
Hungarian (hun) & Original & -- & -- & 3.35 & 36.3 \\
 & Baseline & -- & $1.5\times 10^{-6}$ & 2.76 & 37.4 \\
 & \textbf{+ routing} & 10 & $1.5\times 10^{-6}$ & 2.76 & \textbf{38.1} \\
 & Router-only & 40 & $2\times 10^{-6}$ & 3.34 & 37.1 \\
 & + hidden & 5000 & $8\times 10^{-7}$ & 3.16 & 36.9 \\
\bottomrule
\end{tabular*}
\par\vspace{1em}
\begin{tabular*}{\linewidth}{@{\extracolsep{\fill}}llrrrrrrrrr@{}}
\toprule
\multicolumn{11}{@{}l}{\textbf{(b) Benchmark scores}} \\
\midrule
Lang. & Condition & Belebele & FLORES & \shortstack{MMLU\\Med.} & PIQA & INCLUDE & MGSM & \shortstack{MMLU\\Pro-X} & \shortstack{Multi\\LoKo} & \shortstack{Poly\\Math} \\
\midrule
vie & Original & 73.7 & 30.4 & 52.1 & 68.0 & 52.0 & 50.4 & 35.1 & 10.8 & 25.2 \\
 & Baseline & 73.3 & 38.7 & 52.4 & 72.0 & 49.6 & 54.0 & 33.2 & 9.6 & 27.2 \\
 & \textbf{+ routing} & 72.9 & 38.0 & 52.6 & 72.0 & 51.8 & 50.8 & 33.1 & 10.4 & 29.0 \\
 & Router-only & 73.6 & 32.1 & 51.9 & 66.0 & 51.3 & 55.2 & 35.8 & 11.2 & 26.4 \\
\addlinespace[3pt]
sin & Original & 34.0 & 8.5 & 32.4 & 44.0 & -- & 1.2 & -- & -- & -- \\
 & Baseline & 35.0 & 9.2 & 31.2 & 53.0 & -- & 2.0 & -- & -- & -- \\
 & \textbf{+ routing} & 33.4 & 10.6 & 31.2 & 58.0 & -- & 0.8 & -- & -- & -- \\
 & Router-only & 33.8 & 9.5 & 31.4 & 50.0 & -- & 1.2 & -- & -- & -- \\
 & + hidden & 34.4 & 9.8 & 31.6 & 55.0 & -- & 0.8 & -- & -- & -- \\
\addlinespace[3pt]
hun & Original & 70.6 & 11.7 & -- & 54.0 & 43.3 & 26.4 & 12.0 & -- & -- \\
 & Baseline & 69.3 & 14.8 & -- & 52.0 & 44.0 & 35.2 & 9.2 & -- & -- \\
 & \textbf{+ routing} & 69.0 & 14.0 & -- & 53.0 & 42.5 & 36.8 & 13.2 & -- & -- \\
 & Router-only & 70.3 & 12.3 & -- & 54.0 & 43.3 & 30.0 & 12.4 & -- & -- \\
 & + hidden & 70.6 & 11.9 & -- & 53.0 & 43.6 & 30.4 & 12.1 & -- & -- \\
\bottomrule
\end{tabular*}
\end{table}

%% file: appendix/layervis.tex
\section{Visualizations of MoE Routing For Layer Selection} \label{layervis}

Below, we provide the MoE routing emph{divergence} graphs using the normalized JS-divergence metric defined in \citet{bandarkar2026multilingual}. For Qwen3-30B-A3B and GPT-OSS-20B, that work already selects layers and establishes they work well. Notably, that work identifies that router steering is very sensitive to these exact layers. For Granite-4-H-Tiny and Marco-Nano, we subjectively pick layers based on the below visualizations. The final layer ranges are in the last column of Table~\ref{model_details} in the next section.

\begin{figure}[h]
    \centering
    \includegraphics[width=0.48\textwidth]{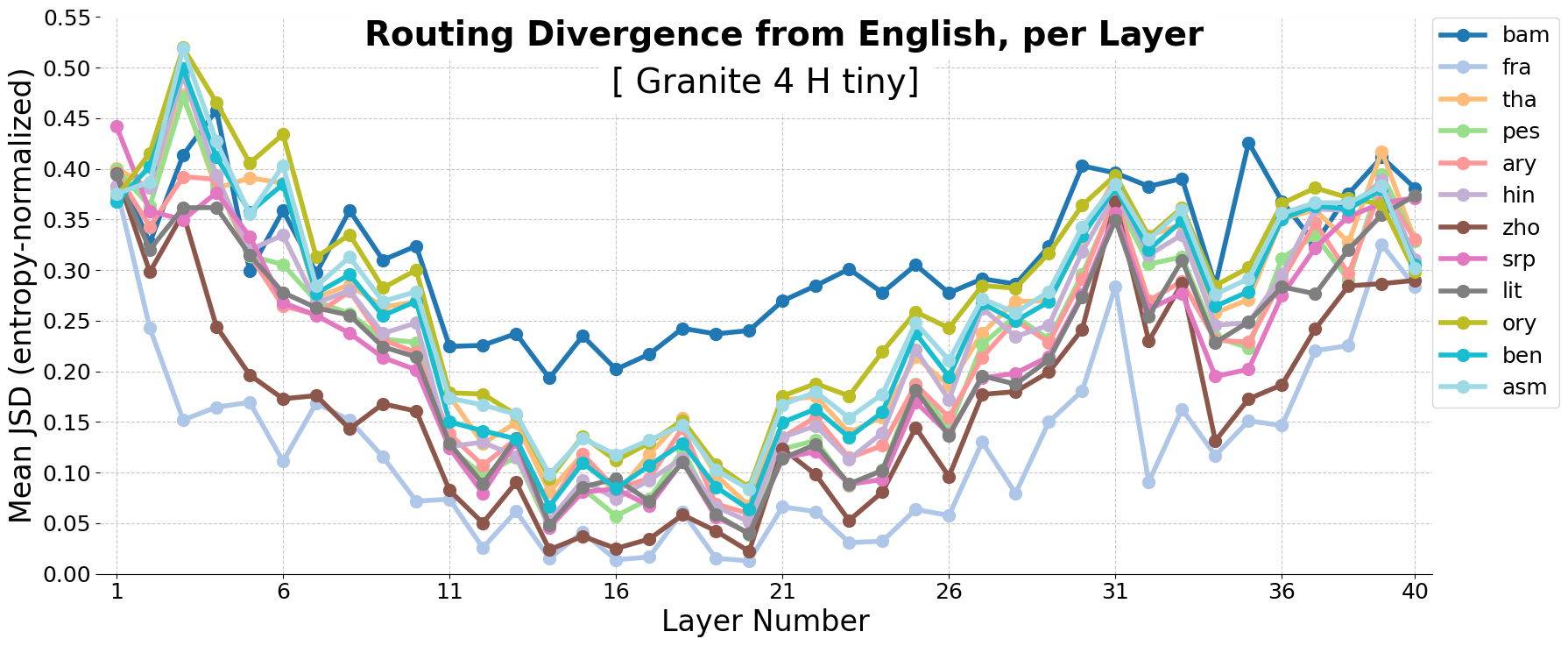}\hfill
    \includegraphics[width=0.48\textwidth]{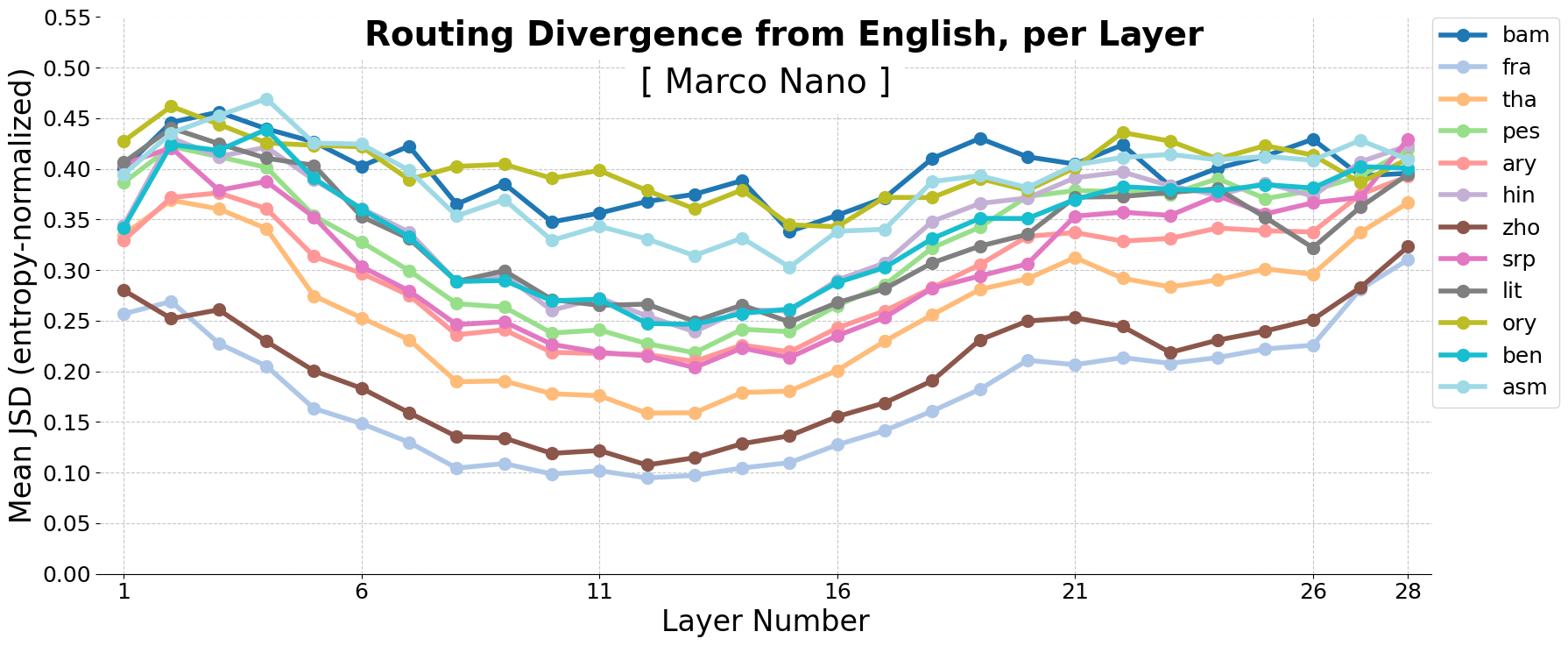}

    \medskip

    \includegraphics[width=0.48\textwidth]{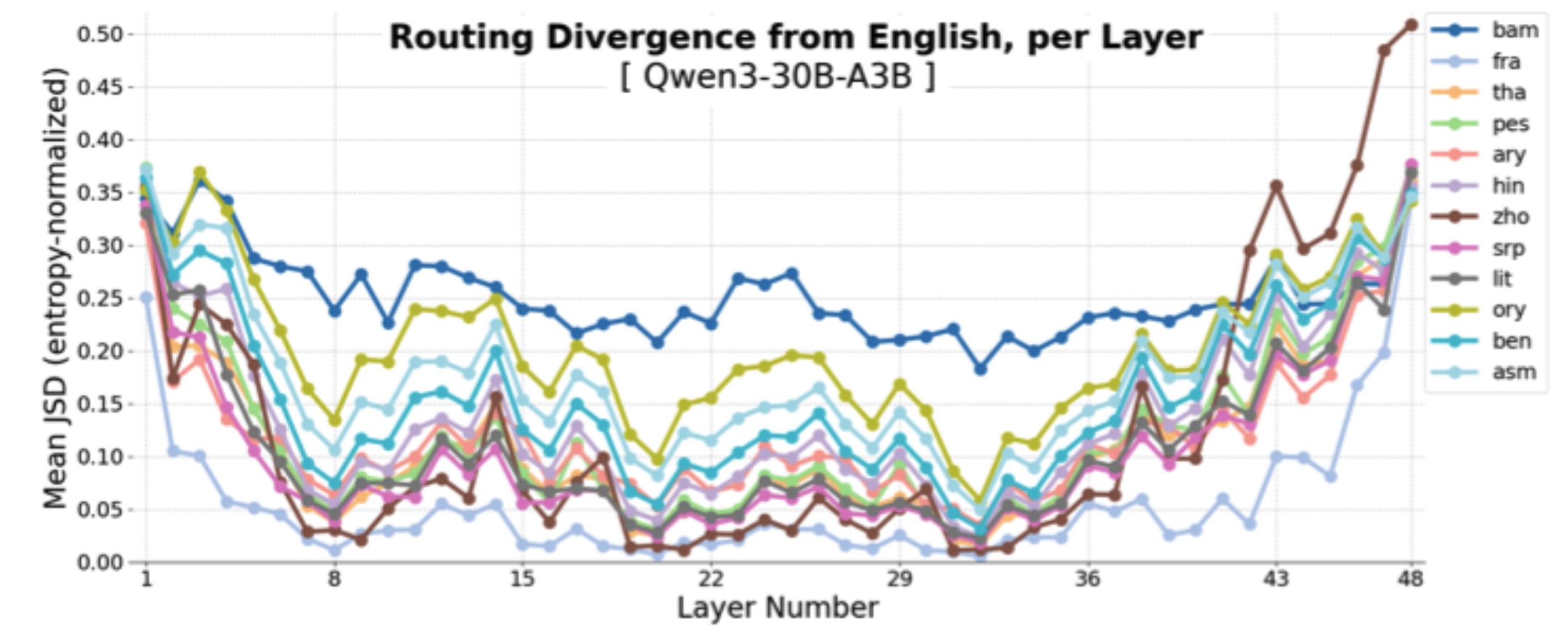}
    \includegraphics[width=0.48\textwidth]{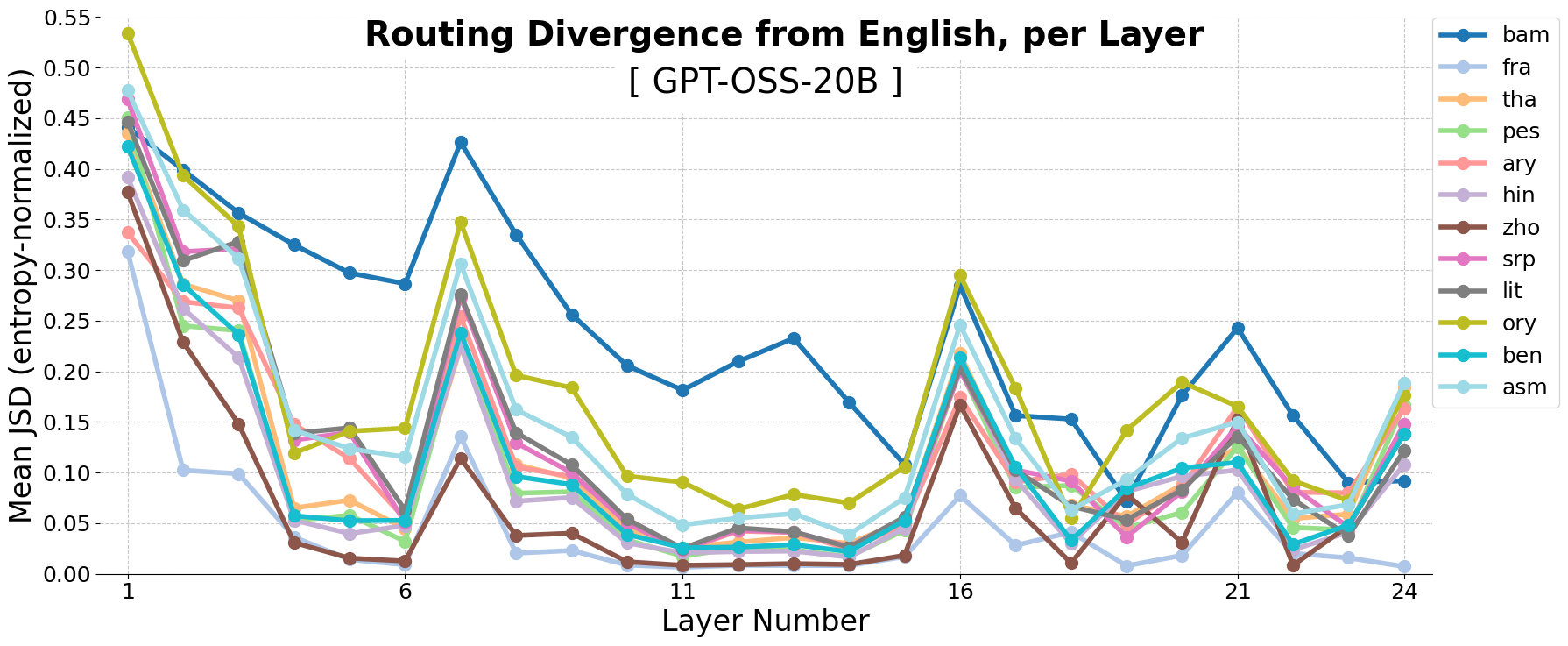}
    \caption{MoE router divergence for the four models used in experiments. Lower values represent higher alignment, and corresponds to higher-resource languages.}
    \label{fig:moe-router-jsd-before-experiments}
\end{figure}

%% file: appendix/model_layers.tex
\section{Model MoE Details}
\label{details}

\begin{table}[H]
    \centering
    \small
    \setlength{\tabcolsep}{3pt}
    \caption{MoE architecture and router-layer details for each model. Active parameter counts are shown in parentheses.}
    \label{model_details}
    \begin{tabular}{@{}lcccc@{}}
        \toprule
        Model & Params (Active) & Num. Layers & Active Experts / Total & Layers for $\mathcal{L}_{XLR}$ \\
        \midrule
        \qwen       & 31B (A3B)   & 48 & 8 / 128 & 7--34 \\
        \gpt         & 22B (A4B)   & 24 & 4 / 32  & 4--17 \\
        \marco          & 8B (A0.6B)  & 28 & 8 / 238 & 7--19 \\
        \granite  & 7B (1B)     & 40 & 6 / 64  & 10--35 \\
        \bottomrule
    \end{tabular}
\end{table}

%% file: appendix/training_data.tex
\section{Parallel Training Data} \label{traindata}

We construct a separate English--target-language parallel dataset for each of the seven languages in our experiments. Each dataset targets 200,000 training examples, with 1,000 additional examples reserved for validation. The examples comprise both sentence-aligned bitext and parallel instruction--response conversations. For instruction-tuning sources, we align the English and target-language versions of the same example and preserve their multi-turn chat structure. We remove empty pairs, pairs whose two sides are identical, and duplicate pairs from the same source.

Table~\ref{tab:training-data-composition} lists the data sources used for each language. The source weights in our data-construction pipeline specify requested allocations rather than the final composition: some sources do not contain enough examples that pass filtering, and their shortfalls are refilled from the remaining sources with available data. Thus, the planned weights do not reliably describe the realized number of examples from each source.

\begin{table}[H]
    \centering
    \small
    \begin{tabular}{p{0.15\linewidth} p{0.77\linewidth}}
        \toprule
        \textbf{Language} & \textbf{Training-data sources} \\
        \midrule
        Vietnamese & MTet \citep{ngo2022mtet}; Bactrian-X \citep{li2023bactrianx}; NLLB English bitext \citep{nllb2022}; OPUS-100 \citep{zhang-etal-2020-improving}. \\
        Sinhala & Bactrian-X \citep{li2023bactrianx}; NLLB English bitext \citep{nllb2022}; OPUS-100 \citep{zhang-etal-2020-improving}. \\
        Hungarian & NLLB English bitext \citep{nllb2022}; OPUS-100 \citep{zhang-etal-2020-improving}. \\
        Telugu & Updesh \citep{chitale-etal-2026-updesh}; Bactrian-X \citep{li2023bactrianx}; Samanantar \citep{ramesh-etal-2022-samanantar}; NLLB English bitext \citep{nllb2022}; OPUS-100 \citep{zhang-etal-2020-improving}. \\
        Kannada & Updesh \citep{chitale-etal-2026-updesh}; Samanantar \citep{ramesh-etal-2022-samanantar}; NLLB English bitext \citep{nllb2022}; OPUS-100 \citep{zhang-etal-2020-improving}. \\
        Thai & Bactrian-X \citep{li2023bactrianx}; SCB-MT-EN-TH-2020 \citep{lowphansirikul2022large}; OPUS-100 \citep{zhang-etal-2020-improving}. \\
        Kyrgyz & NLLB English bitext \citep{nllb2022}; OPUS-100 \citep{zhang-etal-2020-improving}. \\
        \bottomrule
    \end{tabular}
    \caption{Parallel training-data sources for each target language. The table records source inclusion, rather than nominal or realized mixture ratios.}
    \label{tab:training-data-composition}
\end{table}



\paragraph{Length filtering.}
We tokenize both the English and target-language sides before admitting an example and discard examples that exceed the applicable sequence-length limit. The default limit is 768 tokens per side. Because the model tokenizers represent some scripts inefficiently, we use stricter language-specific caps for Sinhala (512 tokens), Kannada (640 tokens), and Telugu (640 tokens). This prevents poorly tokenized examples from producing disproportionately long sequences while retaining the same filtering procedure across sources.

%% file: appendix/hyperparameters.tex
\section{Training Details} \label{hyperparams}

Section~\ref{conditions} describes the training setup and some hyperparameters. All remaining training settings are fixed across experiments. Each run consists of one epoch over 200{,}000 training examples. We use a warmup--stable--decay (WSD) learning-rate schedule, with a linear warmup over the first 10{,}000 examples and a decay over the final 40{,}000 examples. The intervening 150{,}000 examples therefore use a constant learning rate. We optimize with the fused PyTorch implementation of AdamW and do not apply NEFTune. Although the per-device batch size varies with the memory requirements of each model, we adjust the number of gradient-accumulation steps to maintain an effective batch size of 32 examples.

\paragraph{Compute Requirements}
The two larger models require fully sharded data parallelism (FSDP) across multiple GPUs with 80~GB of memory: \gpt{} uses three or four GPUs, depending on the run, while \qwen{} uses four. In contrast, \granite{} and \marco{} each fit on a single GPU. Training \granite{} and \marco{} on one GPU is comparatively slow, with a run taking approximately 15 hours on H100 or A100.